\documentclass[letterpaper, 10 pt, conference]{ieeeconf}  

\IEEEoverridecommandlockouts                              

\usepackage{cite}
\usepackage{amsmath,amssymb,amsfonts}
\usepackage{algorithmic}
\usepackage{graphicx}
\usepackage{textcomp}
\usepackage{xcolor}
\usepackage{mathtools}
\def\BibTeX{{\rm B\kern-.05em{\sc i\kern-.025em b}\kern-.08em
    T\kern-.1667em\lower.7ex\hbox{E}\kern-.125emX}}

\usepackage{subcaption}
\usepackage{url}

\usepackage{eso-pic}

\usepackage{algorithm}

\usepackage[mode=text]{siunitx}
\usepackage{physics}
\AtBeginDocument{\RenewCommandCopy\qty\SI}

\usepackage[hidelinks]{hyperref}
\usepackage{cleveref}
\makeatletter
\let\@ac@firsttrue\relax
\makeatother
\usepackage{acro}
\DeclareAcronym{QP}{
  short = QP,
  long  = quadratic program,
  short-indefinite = a,
  long-indefinite = a
}

\DeclareAcronym{FOC}{
  short = FOC,
  long  = field-oriented control,
  short-indefinite = a,
  long-indefinite = a
}

\DeclareAcronym{ESC}{
  short = ESC,
  long  = electronic speed controller,
  short-indefinite = an,
  long-indefinite = an
}

\DeclareAcronym{LQR}{
  short = LQR,
  long  = linear quadratic regulator,
  short-indefinite = an,
  long-indefinite = a
}

\DeclareAcronym{SITL}{
  short = SITL,
  long  = software-in-the-loop,
  short-indefinite = a,
  long-indefinite = a
}

\DeclareAcronym{HITL}{
  short = HITL,
  long  = hardware-in-the-loop,
  short-indefinite = a,
  long-indefinite = a
}

\DeclareAcronym{FMU}{
  short = FMU,
  long  = flight management unit,
  short-indefinite = a,
  long-indefinite = a
}

\DeclareAcronym{CA}{
  short = CA,
  long  = control allocator,
  short-indefinite = a,
  long-indefinite = a
}

\DeclareAcronym{RTAI}{
  short = RTAI,
  long  = real-time application interface,
  short-indefinite = a,
  long-indefinite = a
}

\DeclareAcronym{DOF}{
  short = DoF,
  long  = degrees-of-freedom,
  short-indefinite = a,
  long-indefinite = a
}

\DeclareAcronym{mocap}{
  short = MoCap,
  long  = motion capture system,
  short-indefinite = a,
  long-indefinite = a
}

\DeclareAcronym{MPC}{
  short = MPC,
  long  = Model Predictive Controller,
  short-indefinite = an,
  long-indefinite = a
}

\DeclareAcronym{NMPC}{
  short = NMPC,
  long  = Nonlinear Model Predictive Control,
  short-indefinite = an,
  long-indefinite = a
}

\DeclareAcronym{PWM}{
  short = PWM,
  long  = pulse width modulation,
  short-indefinite = a,
  long-indefinite = a
}

\DeclareAcronym{EKF}{
  short = EKF,
  long  = extended Kalman filter,
  short-indefinite = an,
  long-indefinite = a
}

\DeclareAcronym{RTOS}{
  short = RTOS,
  long  = real-time operating system,
  short-indefinite = an,
  long-indefinite = a
}

\DeclareAcronym{ISS}{
  short = ISS,
  long  = International Space Station,
  short-indefinite = an,
  long-indefinite = a
}

\DeclareAcronym{FSW}{
  short = FSW,
  long  = flight software,
  short-indefinite = an,
  long-indefinite = a
}

\DeclareAcronym{SFF}{
  short = SFF,
  long  = spacecraft formation flying,
  short-indefinite = an,
  long-indefinite = a
}

\DeclareAcronym{LEO}{
  short = LEO,
  long  = Low-Earth-Orbit,
  short-indefinite = a,
  long-indefinite = a
}

\DeclareAcronym{GNC}{
  short = GNC,
  long  = {Guidance, Navigation \& Control},
  short-indefinite = a,
  long-indefinite = a
}

\DeclareAcronym{ROS2}{
  short = ROS~2,
  long  = Robot Operating System 2,
  short-indefinite = a,
  long-indefinite = a
}

\DeclareAcronym{cFS}{
  short = cFS,
  long  = core Flight System,
  short-indefinite = a,
  long-indefinite = a
}

\DeclareAcronym{ZMQ}{
  short = ZMQ,
  long  = ZeroMQ,
  short-indefinite = a,
  long-indefinite = a
}

\DeclareAcronym{ECI}{
  short = ECI,
  long  = Earth-Centered Inertial,
  short-indefinite = an,
  long-indefinite = a
}

\DeclareAcronym{CW}{
  short = CW,
  long  = Clohessy--Wiltshire,
  short-indefinite = a,
  long-indefinite = a
}

\DeclareAcronym{MRPs}{
  short = MRPs,
  long  = Modified Rodrigues Parameters,
  short-indefinite = an,
  long-indefinite = a
}

\DeclareAcronym{OCP}{
  short = OCP,
  long  = Optimal Control Problem,
  short-indefinite = an,
  long-indefinite = a
}

\DeclareAcronym{SQP}{
  short = SQP,
  long  = Sequential Quadratic Programming,
  short-indefinite = an,
  long-indefinite = a
}

\DeclareAcronym{IMU}{
  short = IMU,
  long  = Inertial Measurement Unit,
  short-indefinite = an,
  long-indefinite = an
}

\crefname{equation}{Eq.}{Eq.}
\Crefname{equation}{Eq.}{Eq.}
\crefname{figure}{Fig.}{Figs.}
\Crefname{figure}{Fig.}{Figs.}
\crefname{section}{Sec.}{Secs.}
\Crefname{section}{Sec.}{Secs.}
\crefname{subsection}{Sec.}{Secs.}
\Crefname{subsection}{Sec.}{Secs.}
\crefname{algorithm}{Algorithm}{Algorithms}

\newcommand{\mat}[1]{\mathbf{#1}}
\newcommand{\vect}[1]{\boldsymbol{#1}}
\newcommand{\quat}[1]{\mathbf{#1}}

\title{\LARGE \bf
Dual--Actuation Emulation of Scaled Relative \\Orbital Dynamics on a Free-Flyer
}

\author{E. Krantz$^{*1}$, P. Roque$^{2}$, G. Tibert$^{1}$, H. Mao$^{1}$, and C. Fuglesang$^{1}$
\thanks{This work was partially supported by the Wallenberg AI, Autonomous Systems and Software Program (WASP) funded by the Knut and Alice Wallenberg Foundation.}
\thanks{$^{*}$Corresponding author: {\tt\small eliaskra@kth.se}}
\thanks{$^{1}$KTH Royal Institute of Technology, Stockholm, Sweden.}
\thanks{$^{2}$California Institute of Technology, Pasadena, United States.}%
}

\begin{document}

\maketitle
\AddToShipoutPictureFG*{\AtPageLowerLeft{%
  \put(\LenToUnit{0.5\paperwidth},\LenToUnit{0.5cm}){\makebox[0pt][c]{%
  \parbox[b]{0.9\paperwidth}{\centering\scriptsize
  \copyright~2026 IEEE. Personal use of this material is permitted. Permission from IEEE
  must be obtained for all other uses, in any current or future media, including
  reprinting/republishing this material for advertising or promotional purposes, creating
  new collective works, for resale or redistribution to servers or lists, or reuse of any
  copyrighted component of this work in other works. Accepted at iSpaRo 2026.}}}}}
\thispagestyle{empty}
\pagestyle{empty}

\begin{abstract}

Ground-based free-flyer testbeds reproduce the planar motion of a free-floating spacecraft using air bearings, but not the relative dynamics of spacecraft in orbit. We present a dual-actuation architecture that physically emulates Clohessy--Wiltshire dynamics on an open-source free-flyer. Two model predictive controllers run in parallel: one drives the solenoid thrusters with an orbital dynamics model, representing the spacecraft under test; the other drives bidirectional propellers that physically apply the orbital accelerations, continuously updating its reference to account for thruster-induced velocity changes. A dynamic similitude maps the orbital scenario onto the limited floor area and actuation rates of the testbed, letting the spacecraft controller run directly at orbital scale. We validate the architecture on hardware and against a high-fidelity orbital simulation, where the orbital accelerations instead arise naturally, both driven by the identical spacecraft controller. Their agreement indicates that the testbed reproduces the relative orbital motion the controller would encounter in flight.

\end{abstract}

\subsection*{Software and Hardware Releases}
\begin{itemize}
    \item Controller stack and simulation:\vspace{0.5mm}\\
    \small{\href{https://github.com/DISCOWER/atmos_cw_emulation}{\nolinkurl{github.com/DISCOWER/atmos_cw_emulation}}}\vspace{1.5mm}
    \item \normalsize Propeller-module documentation:\vspace{0.5mm}\\
    \small{\href{https://atmos.discower.io}{\nolinkurl{atmos.discower.io}}}
\end{itemize}

\section{Introduction}
Ground-based spacecraft simulators are essential tools for developing and validating guidance, navigation, and control algorithms for on-orbit proximity operations~\cite{wilde_historical_2019, rybus_planar_2016}. These testbeds typically employ flat surfaces and air bearings to approximate undamped planar motion, enabling the study of microgravity rendezvous, docking, and formation flying scenarios~\cite{schwartz_historical_2003}. While they closely reproduce free-floating rigid-body dynamics, they do not capture the relative dynamics of spacecraft in orbit, commonly described by the linearized \ac{CW} equations~\cite{clohessy_terminal_1960}. Emulating these dynamics requires physically applying \ac{CW}-equivalent accelerations to the platform while it executes spacecraft maneuvers.

Physically emulating these accelerations on a free-floating testbed has been attempted in only a few works, which fall into two families. One commands the platform's own compressed-air thrusters to apply the \ac{CW}-equivalent accelerations, with the laboratory-to-orbit mapping derived through Buckingham's $\Pi$ theorem, emulating coasting trajectories with an open-loop model-following controller~\cite{ciarcia_emulating_2017}; the same $\Pi$-theorem similitude has since been applied to emulate other scaled space dynamics on air-bearing platforms~\cite{chen_dynamic_2025}. Sharing one actuation system between emulation and spacecraft control has two drawbacks: the on-off, duty-cycle nature of solenoid thrusters limits the resolution of the continuous orbital forces, and a fixed share of the thrust budget must be reserved for the emulation. A second family tilts the support surface so that gravity supplies the in-plane acceleration~\cite{fernandez_development_2023}. Kinematic and hybrid simulators, by contrast, reproduce orbital motion through robotic manipulators or motion tables that constrain the platform~\cite{wilde_historical_2019}.

We present a dual-actuation architecture that emulates relative orbital dynamics on a free-flyer while preserving its free-floating motion, demonstrated on the open-source ATMOS platform~\cite{roque_towards_2025}. ATMOS moves nearly frictionless over a flat, level floor on air bearings, with compressed-air solenoid thrusters providing holonomic three-\ac{DOF} actuation in the plane (position and yaw). By adding a dedicated, propeller-based actuation modality for the emulation, the architecture separates dynamics emulation from spacecraft control, each driven by its own \ac{MPC} and tuned independently (see \cref{fig:architecture}).

Closing the loop on both roles avoids the open-loop error accumulation of model-following emulation and supports active translational and attitude maneuvers concurrently with the applied orbital forces. We map the orbital scenario onto the testbed using the similitude established for \ac{CW} emulation on air-bearing testbeds~\cite{ciarcia_emulating_2017}, applied so that the spacecraft controller runs directly in the orbital scale and is identical on hardware and in high-fidelity simulation. We openly release the controller stack, the propeller-module documentation, and the supporting PX4 airframe, so the emulation can be reproduced on any ATMOS platform and adapted to similar free-flyers.

\subsection*{Contributions}
The main contributions of this work are:
\begin{enumerate}
    \item A dual-actuation architecture that decouples dynamics emulation from spacecraft control, driving each role with its own closed-loop model predictive controller and continuously updating the emulated \ac{CW} acceleration reference to account for the spacecraft's commanded thrust.
    \item A propeller actuation module that applies these accelerations as continuous proportional thrust, removing the duty-cycle quantization of on-off thrusters and adding a second actuation modality to ATMOS.
    \item An experimental and numerical validation comparing hardware emulation on ATMOS against a high-fidelity Basilisk simulation~\cite{kenneally_basilisk_2020}, both driven by the same spacecraft controller operating in the orbital scale.
\end{enumerate}

\section{Scaled Relative Orbital Dynamics}
\label{sec:theory}

\subsection{Relative Orbital Dynamics}
The relative motion of a chaser spacecraft with respect to a target on a circular orbit is described by the \ac{CW} equations~\cite{clohessy_terminal_1960} in the local orbital Hill frame~$\mathcal{H}$. This rotating reference frame is centered at the target, with $\hat{\vect{x}}$ along the radial (outward from the central body), $\hat{\vect{z}}$ along the orbit angular momentum, and $\hat{\vect{y}}$ completing the right-handed frame. The linearized relative equations of motion are:
\begin{equation}
\label{eq:cw}
\begin{aligned}
\ddot{p}_x &= 3n^2 p_x + 2n\dot{p}_y + F_x/m, \\
\ddot{p}_y &= -2n\dot{p}_x + F_y/m, \\
\ddot{p}_z &= -n^2 p_z + F_z/m,
\end{aligned}
\end{equation}
where $p_x, p_y, p_z$ is the relative position of the chaser in $\mathcal{H}$, $m$ its mass, $n = 2\pi/T$ the mean orbital motion, $T$ the orbital period, and $F_x, F_y, F_z$ are external forces acting on the chaser. The radial and along-track ($x$-$y$) dynamics are coupled, whereas the cross-track ($z$) motion is an independent oscillator. We therefore emulate the coupled in-plane dynamics on the planar testbed and restrict the analysis to these two dimensions.

In the absence of external forces, the \ac{CW} equations admit closed-form solutions via the state transition matrix~$\mat{\Phi}(t) \in \mathbb{R}^{4 \times 4}$~\cite{schaub_analytical_2009}:
\begin{equation}
\label{eq:cw_stm}
\begin{bmatrix}
\vect{p}(t) \\
\vect{v}(t)
\end{bmatrix}
= \begin{bmatrix}
\mat{\Phi}_{pp}(t) & \mat{\Phi}_{pv}(t) \\
\mat{\Phi}_{vp}(t) & \mat{\Phi}_{vv}(t)
\end{bmatrix}
\begin{bmatrix}
\vect{p}_0 \\
\vect{v}_0
\end{bmatrix},
\end{equation}
with $c = \cos(nt)$, $s = \sin(nt)$, and
\begin{align}
\mat{\Phi}_{pp} &=\begin{bmatrix}
  4 - 3c & 0 \\ 6(s - nt) & 1
\end{bmatrix},
&\mat{\Phi}_{pv} &= \tfrac{1}{n}\begin{bmatrix}
  s & 2(1{-}c) \\ 2(c{-}1) & 4s {-} 3nt
\end{bmatrix},  \notag \\
\label{eq:stm_entries}
\mat{\Phi}_{vp} &= n\begin{bmatrix}
  3s & 0 \\ 6(c{-}1) & 0
\end{bmatrix},
&\mat{\Phi}_{vv} &= \begin{bmatrix}
  c & 2s \\ -2s & 4c {-} 3
\end{bmatrix},
\end{align}
where $\vect{p} = [p_x,\, p_y]^\top$ is the relative position and $\vect{v} = \dot{\vect{p}}$ the relative velocity in $\mathcal{H}$, with initial conditions $\vect{p}_0 = \vect{p}(0)$ and $\vect{v}_0 = \vect{v}(0)$. The \ac{CW} trajectory is thus fully determined by $n$, $\vect{p}_0$, and $\vect{v}_0$.

\begin{figure}
    \centering
    \includegraphics[width=.7\linewidth]{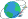}
    \vspace{6pt}
    \caption{Unforced relative motion in the target-centered Hill frame $\mathcal{H}$, showing the along-track drift of a radially offset chaser over successive orbits.}
    \label{fig:hill_frame_drift}
\end{figure}

The coupling between the radial and along-track axes in \cref{eq:cw} means that holding a fixed relative configuration is not a natural solution: left unforced, a chaser displaced from the target drifts away rather than holding formation, as illustrated in \cref{fig:hill_frame_drift}. A bounded, periodic relative orbit instead arises when the initial conditions satisfy the drift-free condition $v_{y,0} = -2n\, p_{x,0}$, which cancels the along-track drift that would otherwise grow without bound. These 2:1 elliptical trajectories, natural solutions of the unforced \ac{CW} equations, take the form
\begin{equation}
\label{eq:stable_ellipse}
\begin{aligned}
p_x(t) &= \alpha \sin(nt + \varphi), \\
p_y(t) &= p_{y,c} + 2\alpha \cos(nt + \varphi),
\end{aligned}
\end{equation}
where $\alpha$ is the semi-minor axis amplitude, $\varphi$ the phase, and $p_{y,c}$ is the along-track center offset.

\subsection{Dynamic Scaling}
\label{sec:scaling}
Orbital maneuvers unfold over tens of minutes to many hours, while the relative trajectories of interest span meters to kilometers, scales far removed from a laboratory experiment of minutes over a floor a few meters across. To reproduce the orbital dynamics within these constraints, we scale the relative motion using the principle of dynamic similitude established for \ac{CW} emulation on air-bearing testbeds~\cite{ciarcia_emulating_2017}, a consequence of the Buckingham $\Pi$ theorem~\cite{buckingham_1914}: when the dimensionless groups that govern two systems take the same values, one is a scaled reproduction of the other.

We define the scale factor of each physical quantity as the ratio of its orbital value to its corresponding laboratory value, the latter marked with a tilde. The scaling is set by four independent factors for time, length, mass, and inertia: $\lambda_t$, $\lambda_l$, $\lambda_m$, and $\lambda_J$. The time and length factors are chosen so that a representative orbital scenario maps onto an experimentally practical duration and the available floor area. The mass and inertia factors are the spacecraft-to-platform mass and inertia ratios. The factor of every other quantity follows from its dimensions: velocity scales as $\lambda_l/\lambda_t$, acceleration as $\lambda_l/\lambda_t^2$, mean motion and angular velocity as $1/\lambda_t$, force as $\lambda_m\lambda_l/\lambda_t^2$, and torque as $\lambda_J/\lambda_t^2$, with actuator bounds scaling like the quantities they constrain.

\begin{figure*}
    \includegraphics[width=\linewidth]{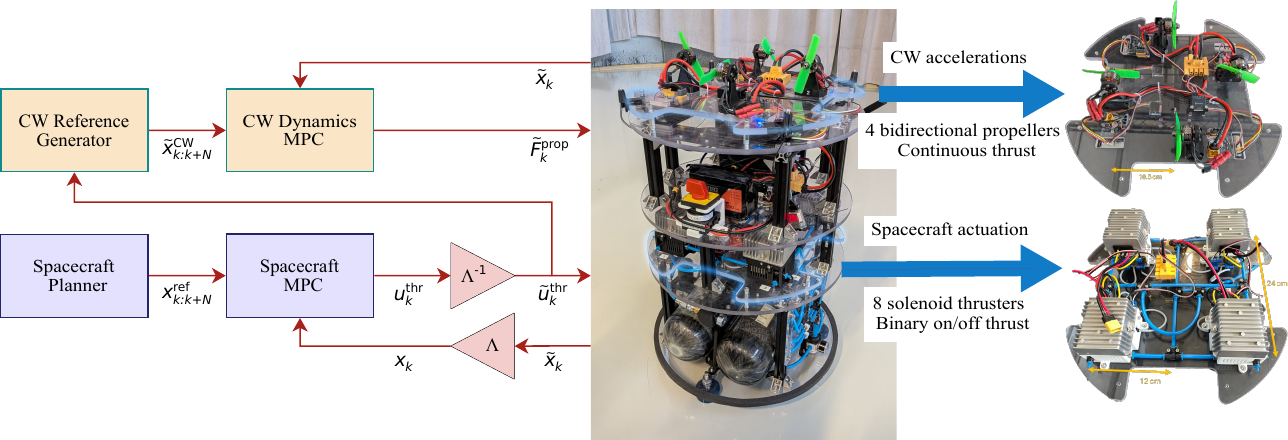}
    \caption{Dual-actuation control architecture. The spacecraft planner and spacecraft \ac{MPC} (orbital scale) generate and track the spacecraft mission trajectory; the commanded thrust updates the \ac{CW} reference generator, which feeds the reference tracked by the \ac{CW} dynamics \ac{MPC} (laboratory scale). $\Lambda$ and $\Lambda^{-1}$ denote the scaling between orbital and laboratory frames. Subscript $k{:}k{+}N$ denotes a trajectory over the prediction horizon and subscript $k$ a single value at time step $k$. The spacecraft \ac{MPC} commands the solenoid thrusters, while the \ac{CW} dynamics \ac{MPC} drives the bidirectional propellers to impose the scaled relative dynamics. The ATMOS free-flyer is shown equipped with the new propeller module.}
    \label{fig:architecture}
\end{figure*}

Substituting these factors into \cref{eq:cw} leaves the in-plane \ac{CW} equations invariant in the laboratory variables,
\begin{equation}
\label{eq:cw_scaled}
\begin{aligned}
\ddot{\tilde{p}}_x &= 3\tilde{n}^2 \tilde{p}_x + 2\tilde{n}\dot{\tilde{p}}_y + \tilde{F}_x/\tilde{m}, \\
\ddot{\tilde{p}}_y &= -2\tilde{n}\dot{\tilde{p}}_x + \tilde{F}_y/\tilde{m},
\end{aligned}
\end{equation}
with laboratory mean motion $\tilde{n} = \lambda_t\, n$. The equations retain their orbital form, so the scaling preserves the \ac{CW} dynamics exactly rather than approximating them. Inertia does not enter these translational equations, so the similitude of \cite{ciarcia_emulating_2017} requires no inertia factor. It is needed here only because the free-flyer carries a physical moment of inertia that couples the thruster torques to its angular acceleration, and through this rotational coupling, the platform reproduces the attitude maneuvers of the represented spacecraft. The rigid-body relation $\tilde{J}\dot{\tilde{\omega}} = \tilde{\tau}$ is then the scaled image of $J\dot{\omega} = \tau$, so attitude maneuvers are reproduced at rates consistent with the scaled translational motion.

\section{Dual--Actuation Control Architecture}
\label{sec:control}

The architecture pairs the free-flyer's two actuation modalities with two \acp{MPC} running in parallel, linked by a \ac{CW} reference generator (\cref{fig:architecture}). The spacecraft \ac{MPC} represents the spacecraft's onboard controller and is the system under test: it carries the \ac{CW} dynamics in its prediction model and runs in the orbital scale, operating under the relative-orbital dynamics it would experience in flight. The \ac{CW} dynamics \ac{MPC} drives the bidirectional propellers to reproduce the scaled relative dynamics of \cref{eq:cw_scaled} on the free-flyer, and therefore runs in the laboratory scale; the propeller modality only imposes those dynamics, leaving the spacecraft \ac{MPC} unchanged. The \ac{CW} reference generator turns the thrust commanded by the spacecraft \ac{MPC} into the relative-orbit trajectory that the \ac{CW} dynamics \ac{MPC} tracks. Because the spacecraft \ac{MPC} runs in the orbital scale, the identical controller can be validated on the free-flyer and in high-fidelity simulation without modification.

\subsection{Nonlinear Model Predictive Control}

Both controllers are formulated as nonlinear \ac{MPC} optimization problems~\cite{rawlings_model_2020}. At each discrete time~$k$, each solves the following \ac{OCP} over a horizon of $N$~steps:
\begin{equation}
\label{eq:nmpc_ocp}
\begin{aligned}
J^*(\vect{x}_k) &= \min_{\vect{u}_k} \sum_{i=0}^{N-1} \ell\big(\vect{x}(i|k), \vect{u}(i|k), \vect{x}^{\text{ref}}(i|k)\big) \\
&\hphantom{{}= \min_{\vect{u}_k} \sum_{i=0}^{N-1}} +\ell_N\big(\vect{x}(N|k), \vect{x}^{\text{ref}}(N|k)\big) \\
\text{s.t.} \quad & \vect{x}(0|k) = \vect{x}_k, \\
& \vect{x}(i+1|k) = f\big(\vect{x}(i|k), \vect{u}(i|k)\big), \\
& \vect{x}(i|k) \in \mathbb{X},\ i = 0, \dots, N, \\
& \vect{u}(i|k) \in \mathbb{U},\ i = 0, \dots, N{-}1.
\end{aligned}
\end{equation}
The stage cost $\ell(\cdot)$ and terminal cost $\ell_N(\cdot)$ are weighted quadratic penalties on the tracking error with respect to the reference $\vect{x}^{\text{ref}}(i|k)$, supplied over the full horizon, with the stage cost also penalizing the control effort; the specific terms of each controller are given below. The terminal cost approximates the cost-to-go beyond the finite horizon by weighting the terminal tracking error more heavily than the stage cost, and replaces a terminal constraint: the terminal state is constrained only to the state set $\mathbb{X}$, so the reference enters the problem only through the cost, and the OCP remains feasible for any reference, including step changes. Closed-loop convergence therefore relies on the terminal cost and horizon length rather than on a terminal region.

\subsection{Spacecraft MPC}
\label{sec:thruster_mpc}

The spacecraft \ac{MPC} represents the spacecraft's onboard controller. Its prediction model incorporates the \ac{CW} accelerations, so that it predicts and compensates for the orbital environment it believes itself to be operating in. With state $\vect{x} = [\vect{p}, \vect{v}, \quat{q}, \vect{\omega}]$ of relative position, velocity, attitude quaternion (body to Hill frame), and body-frame angular velocity, the continuous-time dynamics are:
\begin{equation}
\label{eq:dyn_thruster}
\begin{aligned}
\dot{\vect{p}} &= \vect{v}, &
\dot{\vect{v}} &= \vect{a}^\mathrm{CW}(\vect{p}, \vect{v}) + \tfrac{1}{m}\, \mat{R}(\quat{q})\, \vect{F}^\mathrm{thr}, \\
\dot{\quat{q}} &= \tfrac{1}{2}\, \mat{\Omega}(\vect{\omega})\, \quat{q}, &
\dot{\vect{\omega}} &= \mat{J}^{-1}\!\left(\vect{\tau}^\mathrm{thr} - \vect{\omega} \times \mat{J}\vect{\omega}\right),
\end{aligned}
\end{equation}
where $\vect{a}^\mathrm{CW}(\vect{p}, \vect{v})$ is the \ac{CW} acceleration of~\cref{eq:cw}, $\mat{R}(\quat{q})$ is the rotation matrix from body to Hill frame, $\vect{u}^\mathrm{thr} = [{\vect{F}^\mathrm{thr}},\, {\vect{\tau}^\mathrm{thr}}]$ is the commanded body-frame force and torque, $\mat{\Omega}(\vect{\omega})$ is the quaternion kinematic matrix, and $\mat{J}$ the spacecraft inertia matrix.
The stage cost penalizes the position, velocity, attitude, and angular-velocity tracking errors and the control effort, where the attitude error is the vector part of the error quaternion $\quat{q}_\mathrm{e}=\quat{q}^{\mathrm{ref}\,*} \otimes \quat{q}$ and the remaining errors are differences. The reference $\vect{x}^\mathrm{ref}(i|k)$ is a preplanned spacecraft mission trajectory. The spacecraft \ac{MPC} is unaware of the propeller module and controller.

\subsection{CW Dynamics MPC}
\label{sec:prop_mpc}

The \ac{CW} dynamics \ac{MPC} physically applies the \ac{CW} accelerations to the free-flyer with the four propellers, using free-floating rigid-body dynamics. Let $\vect{F}^\mathrm{prop} = [F^\mathrm{prop}_1,\, \dots\,,\, F^\mathrm{prop}_4] \in \mathbb{R}^4$ denote the four propeller forces. To respect the propellers' finite thrust slew rate, these forces are carried in the state, $\vect{x} = [\vect{p},\, \vect{v},\, \quat{q},\, \vect{\omega},\, \vect{F}^\mathrm{prop}]$, and the control input is their rate $\vect{u}^\mathrm{prop} = \dot{\vect{F}}^\mathrm{prop}$, so that a hard bound $|\vect{u}^\mathrm{prop}| \le \dot{F}^\mathrm{prop}_\mathrm{max}$ limits how fast the commanded force can change. As this controller runs at laboratory scale, all quantities are laboratory values. The dynamics are
\begin{equation}
\label{eq:dyn_prop}
\begin{aligned}
\dot{\vect{p}} &= \vect{v}, &
\dot{\vect{v}} &= \tfrac{1}{m}\, \mat{R}(\quat{q})\, \mat{B}_F\, \vect{F}^\mathrm{prop}, \\
\dot{\quat{q}} &= \tfrac{1}{2}\, \mat{\Omega}(\vect{\omega})\, \quat{q}, &
\dot{\vect{\omega}} &= \mat{J}^{-1}\!\left(\mat{B}_\tau\, \vect{F}^\mathrm{prop} - \vect{\omega} \times \mat{J}\vect{\omega}\right), \\
\dot{\vect{F}}&^\mathrm{prop} = \vect{u}^\mathrm{prop},
\end{aligned}
\end{equation}
where $\mat{B}_F,\, \mat{B}_\tau \in \mathbb{R}^{3\times4}$ map the propeller forces to the body-frame force and torque, and $\vect{\tau}^\mathrm{prop} = \mat{B}_\tau \vect{F}^\mathrm{prop}$.

The stage cost is
\begin{equation}
\label{eq:prop_cost}
\ell(\cdot) =  \|\vect{e}_p\|^2_{\mat{Q}_p}
  + \|\vect{e}_v\|^2_{\mat{Q}_v}
+ \|\tau_z^\mathrm{prop}\|^2_{Q_\tau}
  + \|\vect{F}^\mathrm{prop}\|^2_{\mat{Q}_F},
\end{equation}
where $\vect{e}_p$ and $\vect{e}_v$ are the position and velocity tracking errors with respect to the \ac{CW} reference of~\cref{sec:cw_refgen}, $\tau_z^\mathrm{prop}$ the net yaw torque, and $\|\vect{F}^\mathrm{prop}\|^2_{\mat{Q}_F}$ the control-effort term. The force rate is left unpenalized and is instead regularized by the solver.

The controller does not track attitude, since the \ac{CW} dynamics are purely translational, and the large weight $Q_\tau$ on the net yaw torque keeps the propellers from rotating the free-flyer, leaving attitude maneuvers to the spacecraft \ac{MPC}. Even so, $\quat{q}$ and $\vect{\omega}$ are required in the state: $\mat{R}(\quat{q})$ maps the body-frame propeller force into the Hill frame, and this mapping changes as the free-flyer yaws.

\subsection{CW Reference Generator}
\label{sec:cw_refgen}
 
The reference $\tilde{\vect{x}}_{k{:}k{+}N}^\mathrm{CW}$ for the \ac{CW} dynamics \ac{MPC} is a full trajectory over the prediction horizon, computed analytically in the laboratory scale from the \ac{CW} state transition matrix of~\cref{eq:cw_stm} with mean motion~$\tilde{n}$ (\cref{alg:cw_refgen}). The generator is initialized at the measured position with zero velocity, so the free-flyer is first brought close to rest before the emulation starts. As the spacecraft \ac{MPC} maneuvers the spacecraft, the initial conditions of $\mat{\Phi}(t)$ are updated to reflect the commanded thrust. At each update, the generator propagates the current initial conditions to the present time, adds the velocity change imparted by the thrust held since the previous update, and resets the epoch. The updated initial conditions define a new \ac{CW} reference, ensuring continuity.
 
\begin{algorithm}[t]
\caption{CW reference generator}
\label{alg:cw_refgen}
\begin{algorithmic}[1]
\REQUIRE $\tilde{n}$, $\tilde{m}$, step $T_\mathrm{s}$, horizon $N$
\STATE $\tilde{\vect{p}}_0 \gets$ measured position, $\tilde{\vect{v}}_0 \gets \vect{0}$, $\tilde{\vect{F}} \gets \vect{0}$, $t_0 \gets t$
\item[] \textit{Update (periodically and on new $\tilde{\vect{F}}^\mathrm{thr}$):}
\STATE $\Delta t \gets t - t_0$
\STATE $(\tilde{\vect{p}}, \tilde{\vect{v}}) \gets \mat{\Phi}(\Delta t)\,(\tilde{\vect{p}}_0, \tilde{\vect{v}}_0)$
\STATE $\Delta\tilde{\vect{v}} \gets \dfrac{1}{\tilde{m}}\tilde{\vect{F}}\,\Delta t$
\STATE $\tilde{\vect{p}}_0 \gets \tilde{\vect{p}}$, \
       $\tilde{\vect{v}}_0 \gets \tilde{\vect{v}} + \Delta\tilde{\vect{v}}$, \
       $t_0 \gets t$
\IF{new command $\tilde{\vect{F}}^\mathrm{thr}$ received}
    \STATE $\tilde{\vect{F}} \gets \mat{R}(\tilde{\quat{q}})\,\tilde{\vect{F}}^\mathrm{thr}$
\ENDIF
\item[] \textit{Output at time step $k$:}
\FOR{$i = 0, \dots, N$}
    \STATE $\tilde{\vect{x}}_{k+i}^\mathrm{CW} \gets
        \mat{\Phi}(t_k - t_0 + i\,T_\mathrm{s})\,(\tilde{\vect{p}}_0, \tilde{\vect{v}}_0)$
\ENDFOR
\STATE send $\tilde{\vect{x}}^\mathrm{CW}_{k:k+N}$ to the \ac{CW} dynamics \ac{MPC}
\end{algorithmic}
\end{algorithm}

This feedback of the commanded thrust into the reference closes the loop between the two controllers: rather than replaying a precomputed force history open-loop, the generator regenerates the relative-orbit reference online as the spacecraft maneuvers. Because the \ac{CW} dynamics \ac{MPC} receives a full predicted trajectory rather than a single setpoint, it anticipates the orbital motion and tracks it proactively, giving smoother force profiles and reduced tracking lag.

\subsection{Numerical Normalization}

Unlike the laboratory scale, where the state and input are naturally well-conditioned, the orbital scale spans many orders of magnitude: the relative position can be large while the relative velocities and the corresponding forces are comparatively small, so the \ac{OCP} becomes numerically ill-conditioned. We therefore normalize the states, references, and inputs by their characteristic values before passing them to the solver and denormalize the predicted solution afterward. The cost is evaluated in normalized coordinates, so the weight matrices are tuned for the normalized states and inputs.
\section{Validation}
\label{sec:experiment}

We validate the architecture by running one scenario in two settings: physically emulated on the ATMOS free-flyer, and reproduced in the high-fidelity Basilisk simulator. The same spacecraft controller drives both, so agreement between them confirms that the emulation reproduces the orbital dynamics the controller would experience in flight.

\subsection{Emulated Scenario}
We emulate a chaser near a passive target on a circular low-Earth orbit with a period of \qty{100}{\minute}. The represented spacecraft has a mass of \qty{152}{\kg} and a yaw inertia of \qty{24}{\kg\metre\squared}; against the \qty{19}{\kg}, \qty{0.3}{\kg\metre\squared} free-flyer this fixes the mass and inertia factors \(\lambda_m = 8\) and \(\lambda_J = 80\). We select a time factor \(\lambda_t = 50\) and length factor \(\lambda_l = 100\), mapping the orbit onto a \qty{2}{\minute} laboratory period while keeping the relative motion within the $\sim$\qty{15}{\metre\squared} floor.

The chaser first holds three static waypoints in the target's Hill frame, at \((p_x, p_y) = (50, 100)\), \((-50, 150)\), and \((50, 200)\)\,\unit{\m} in orbital scale, performing a \ang{45} yaw maneuver between consecutive waypoints. None of these is a natural \ac{CW} solution, so the chaser would drift away without continuous station-keeping thrust. Each waypoint is supplied to the spacecraft \ac{MPC} as a constant reference over the horizon and switched to the next as a step, without intermediate points. It then transfers to two stable 2:1 relative orbits of the form in \cref{eq:stable_ellipse}, supplied as the analytic trajectory sampled at the controller's time steps, both centered at \(p_{y,c} = \qty{200}{\m}\) with semi-minor axes \(\alpha = \qty{50}{\m}\) and \qty{37.5}{\m}. On these orbits, the planned trajectory is dynamically consistent with the applied \ac{CW} accelerations, so the spacecraft controller should require almost no thrust.

\begin{figure}[t]
    \centering
    \includegraphics[height=6.7cm]{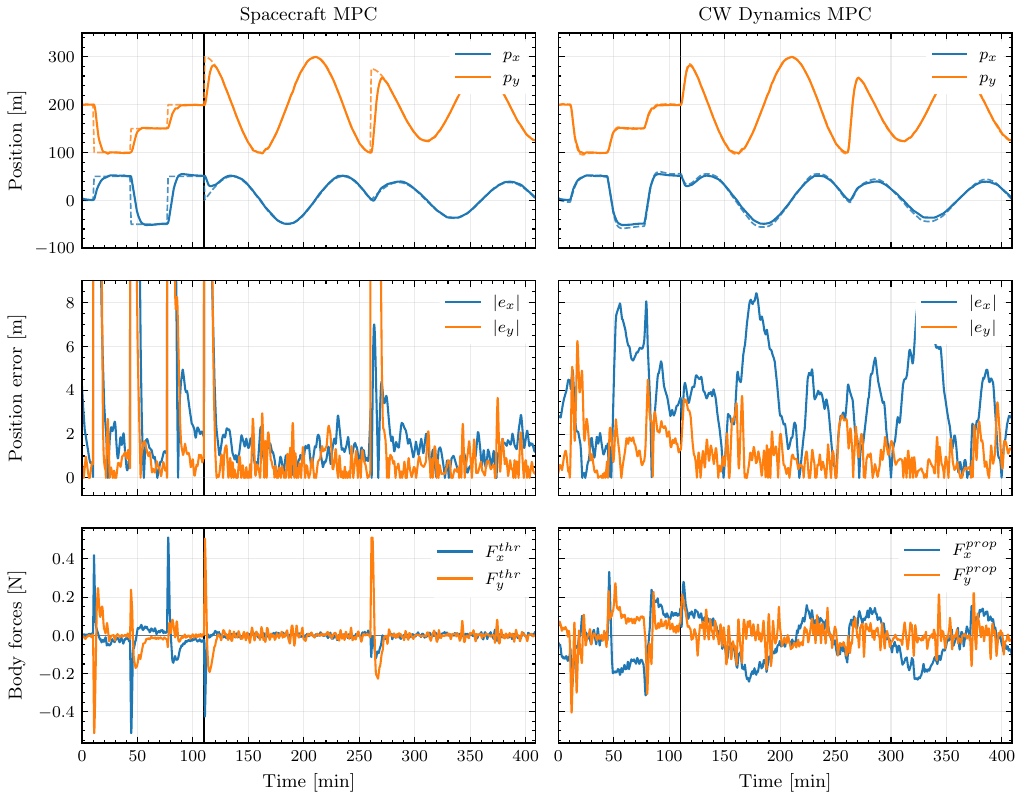}
    \caption{Hardware results on ATMOS, shown at orbital scale (\(\lambda_t=50\), \(\lambda_l=100\), \(\lambda_F=\lambda_m\lambda_l/\lambda_t^2=0.32\)). Left column: spacecraft \ac{MPC}; right column: \ac{CW} dynamics \ac{MPC}. The chaser first holds three unstable static formations relative to the target, and then at $t\geq$ \qty{110}{\minute} follows two consecutive stable relative orbits. Top: measured (solid) and reference (dashed) radial \(p_x\) and along-track \(p_y\) positions. Middle: position tracking error. Bottom: commanded body forces. In the stable orbits, the spacecraft thrust nearly vanishes while the propellers supply the continuous \ac{CW} accelerations.}
    \label{fig:translation_plot_atmos}
\end{figure}

\begin{figure}[t]
    \hspace{1.4cm}
    \includegraphics[height=6.7cm]{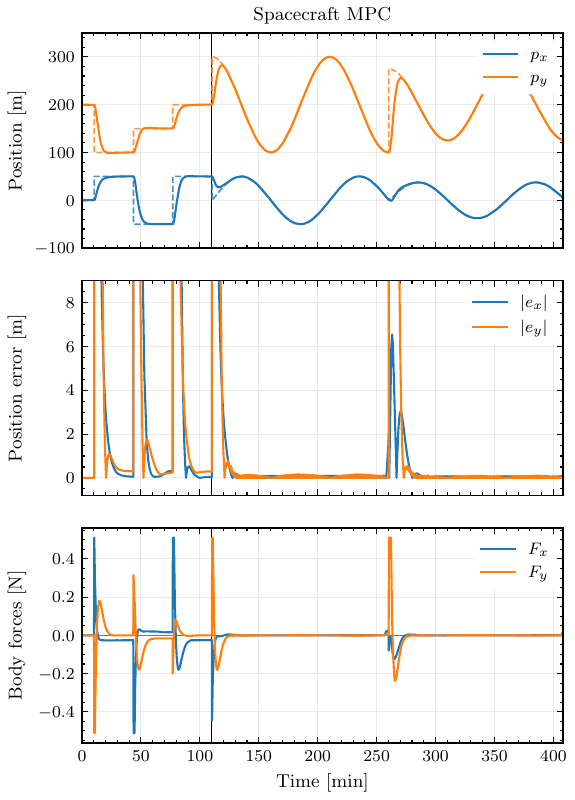}
    \caption{Basilisk reference results, at the native orbital scale. Only the spacecraft \ac{MPC} runs since the nonlinear orbital dynamics of Basilisk replace the \ac{CW} dynamics \ac{MPC}. Without modeled disturbances, the controller tracks the reference closely and coasts on the stable orbits ($t\geq$ \qty{110}{\minute}).}
    \label{fig:translation_plot_bsk}
\end{figure}

\subsection{Implementation}
The hardware experiments run on a single ATMOS free-flyer~\cite{roque_towards_2025} carrying both actuation modules and floating on air bearings over a flat, leveled floor. Eight compressed-air solenoid thrusters provide holonomic three-\ac{DOF} actuation, delivering up to about \qty{3}{\newton} of force and \qty{0.5}{\newton\meter} of yaw torque in short \qty{10}{\hertz} pulses. The propeller module adds four bidirectional brushless motors, offset \qty{10.5}{\centi\meter} from the center of mass, with field-oriented speed controllers, supplying continuous proportional thrust up to about \qty{4}{\newton} of force and \qty{0.8}{\newton\metre} of torque. Both controllers run in \ac{ROS2}~\cite{ros2paper} on a laboratory computer (Intel Core i7-12700H, Ubuntu 22.04), which communicates with the free-flyer over a wireless network. Onboard, a PX4~\cite{meier_px4_2015} flight controller communicates over micro XRCE-DDS via an NVIDIA Jetson, handling sensor data and low-level actuation, commanding the solenoid thrusters by PWM and the propeller speed controllers over CAN. Because the laboratory runs \(\lambda_t = 50\) times faster than the orbital scenario, all controller and actuator rates are \(\lambda_t\) times their orbital-scale values. Both controllers are solved with Acados~\cite{verschueren_acadosmodular_2022} using a real-time iteration sequential quadratic programming scheme with an HPIPM \ac{QP} solver, over a horizon of \(N = 20\) steps of \qty{0.2}{\second}, that is \qty{4}{\second} in the laboratory and \qty{200}{\second} at orbital scale. The spacecraft \ac{MPC} runs at \qty{10}{\hertz} (\qty{0.2}{\hertz} orbital), matching the thruster pulse rate, and the \ac{CW} dynamics \ac{MPC} at \qty{20}{\hertz}. The mean (maximum) solve times are \qty{1.4}{\milli\second} (\qty{2.4}{\milli\second}) for the spacecraft \ac{MPC} and \qty{1.1}{\milli\second} (\qty{2.8}{\milli\second}) for the \ac{CW} dynamics \ac{MPC}, well below the respective \qty{100}{\milli\second} and \qty{50}{\milli\second} control periods, so the rates are set by the actuation rather than by computation.

As a high-fidelity reference, we reproduce the same scenario in the Basilisk astrodynamics simulator~\cite{kenneally_basilisk_2020}, which integrates the full nonlinear orbital dynamics rather than their \ac{CW} linearization, with the represented spacecraft on the \qty{100}{\minute} orbit and the target at the Hill-frame origin. Through the Basilisk-\ac{ROS2} bridge~\cite{krantz2025bridging}, the identical \ac{ROS2} spacecraft controller, including its weighting matrices, runs against the simulator, with only the PX4 interface messages replaced by their Basilisk equivalents. Since the orbital accelerations arise naturally in the simulator, no \ac{CW} dynamics \ac{MPC} is required and the spacecraft controller runs alone.

\subsection{Results}
\cref{fig:translation_plot_atmos} shows the hardware results and \cref{fig:translation_plot_bsk} the Basilisk reference, both at orbital scale, with the laboratory measurements scaled up accordingly. On the hardware, the spacecraft \ac{MPC} tracks the planned trajectory throughout. While holding the unstable waypoints, it thrusts continuously to oppose the \ac{CW} accelerations; once on the stable relative orbits its thrust nearly vanishes, with only occasional corrections below \qty{50}{\milli\newton} (\qty{160}{\milli\newton} in lab scale), while the \ac{CW} dynamics \ac{MPC} supplies the continuous force that realizes the orbital accelerations, bounded by about \qty{230}{\milli\newton} (\qty{720}{\milli\newton} in lab scale). This separation emerges even though the \ac{CW} reference generator receives only the commanded thrust and never the spacecraft controller's plan, confirming that the propellers impose the expected accelerations independently.
The spacecraft \ac{MPC} also performs the \ang{45} yaw maneuvers between waypoints, keeping the attitude error below \ang{6} while the propellers hold their commanded net torque near zero.

Once converged on the stable orbits, the spacecraft \ac{MPC} tracking error stays below \qty{3}{\m} (\qty{3}{\cm} in lab scale), and the \ac{CW} dynamics \ac{MPC} tracks its \ac{CW} reference to within \qty{8}{\m} (\qty{8}{\cm}).

In Basilisk, the same spacecraft controller tracks the planned trajectory to within \qty{20}{\cm}. With no disturbances or sensor noise, the error is far smaller than on hardware, and the thrust falls essentially to zero on the stable orbits. The close agreement in controller behavior across the physical and numerical settings, produced by a single unmodified controller, is the validation we set out to demonstrate.
\section{Discussion}
\label{sec:discussion}

The experiments confirm the intended decoupling: the spacecraft \ac{MPC} tracks its reference while the propellers independently impose the scaled \ac{CW} accelerations, and the two roles interact only through the commanded thrust. The spacecraft controller also performs the yaw maneuvers concurrently with the translational emulation. Penalizing the net propeller torque keeps the propellers from inducing a parasitic torque on the platform, leaving the attitude free for the spacecraft controller to maneuver. Carrying the propeller force as a state with a hard slew-rate limit (\cref{sec:prop_mpc}) further respects the actuators' finite response and smooths the applied force.

Prior thruster-based emulation relied on open-loop model following, where position and velocity errors accumulate over a finite-duration maneuver and degrade the emulated trajectory~\cite{ciarcia_emulating_2017}; closing the loop on the emulation removes this error-accumulation path, since the \ac{CW} dynamics \ac{MPC} tracks a continuously regenerated reference rather than replaying a precomputed force history. The residual tracking errors are believed to stem mainly from imperfect frictionless support and from model mismatch, neither of which the present controllers estimate. An offset-free \ac{MPC} with online disturbance estimation~\cite{morari_nonlinear_2012} would absorb much of this residual and is a natural next step. Closing the loop also means that the reference is built from the commanded thrust, since the thrust delivered by compressed-air thrusters can neither be modeled precisely nor measured directly. Any difference between the two appears as a tracking error that the propellers reject, so the emulated spacecraft behaves largely as if its thrusters delivered the commanded force.

Operating the controllers natively in the orbital scale raised a conditioning issue. There, the state and input could span many orders of magnitude, since the relative positions grow with the length scale while the attitude errors, angular rates, and applied forces remain small. The solver converged reliably only after states and inputs were normalized by their characteristic values, a step we found essential to running \ac{MPC} directly in the orbital scale and one that is distinct from the physical scaling of the dynamics.

The approach is not restricted to circular orbits. The same architecture can carry a higher-fidelity relative model in the controllers, such as the elliptical-orbit generalization of the \ac{CW} equations~\cite{yamanaka_ankersen_2002}, to emulate eccentric-orbit rendezvous. The planar floor omits the independent cross-track \ac{CW} mode, which could be emulated by projecting the out-of-plane motion onto the plane. The scaled forces must also remain within the actuation envelope of both modalities, the continuous propeller thrust and the peak solenoid thrust, which bounds the admissible mass, length and time scales, as well as the represented spacecraft thrust. This was not limiting in our experiments but could constrain more aggressive maneuvers.

A natural next step is to emulate full rendezvous and docking under these scaled \ac{CW} dynamics, testing a spacecraft controller through approach and capture on the free-flyer rather than against waypoints and relative orbits alone. Finally, the propeller module and its controller are not specific to \ac{CW} emulation. The same continuous, proportional actuation can emulate the dynamics of intravehicular free-flyers such as Astrobee~\cite{bualat_astrobee_2015} or Int-Ball~\cite{mitani_intball_2019}, which use proportional fans rather than binary valves, and can apply arbitrary programmed forces, whether to inject controlled disturbances or to actively cancel unwanted ones.
\section{Conclusion}

We presented a dual-actuation architecture that physically emulates relative orbital dynamics on a planar free-flyer while preserving its free-floating motion. Separating spacecraft control (solenoid thrusters) from dynamics emulation (bidirectional propellers) into two parallel model predictive controllers decouples controller design from environment emulation at the actuation level, and a continuously updated \ac{CW} reference generator links the two without coupling their controllers. On the ATMOS free-flyer, the spacecraft controller tracked scaled \ac{CW} trajectories, including concurrent attitude maneuvers, while the propellers supplied the continuous orbital forces, and the identical controller reproduced the same behavior against a high-fidelity Basilisk simulation. The approach turns a planar air-bearing testbed into a physical emulator of orbital relative motion, and because each free-flyer emulates its own dynamics, it invites extensions to eccentric orbits, full rendezvous and docking, and multiple independently emulated agents.

\bibliographystyle{IEEEtran}
\bibliography{references/eliaskra_ref}

\end{document}